\documentclass{article} 
\usepackage{iclr2027_conference,times}

\usepackage{amsmath,amsfonts,bm}

\def\eqref#1{equation~\ref{#1}}

\def\1{\bm{1}}

\DeclareMathAlphabet{\mathsfit}{\encodingdefault}{\sfdefault}{m}{sl}
\SetMathAlphabet{\mathsfit}{bold}{\encodingdefault}{\sfdefault}{bx}{n}

\usepackage{hyperref}
\hypersetup{
    colorlinks,
    linkcolor={red!50!black},
    citecolor={blue!50!black},
    urlcolor={blue!80!black}
}
\usepackage{url}
\usepackage{booktabs}       
\usepackage{amsfonts}       
\usepackage{nicefrac}       
\usepackage{microtype}      
\usepackage{xcolor}         
\usepackage{adjustbox}
\usepackage{xspace}
\usepackage{amsmath}
\usepackage{mathtools}
\usepackage{amssymb}
\usepackage{graphicx}
\usepackage{wrapfig}
\usepackage[table]{xcolor} 
\usepackage{tcolorbox}
\usepackage{rotating}
\usepackage{bbm}
\usepackage{tabularx}
\usepackage{longtable}

\tcbuselibrary{skins}

\usepackage{pifont}
\usepackage{booktabs,makecell}

\title{OSWorld-Pro: Process-based Evaluation for Computer Use Agents}

\author{Zhilin Wang, Shaokun Zhang, Yifan Zhang, Hao Zhang, Jin Xu, Binfeng Xu,\\
\textbf{Jian Hu, Yunheng Zou, Karan Sapra, Andrew Tao, Jan Kautz, Yi Dong}  \\
NVIDIA\\
\texttt{\{zhilinw, yidong\}@nvidia.com} \\
}

\iclrfinalcopy 
\begin{document}

\maketitle

\begin{abstract}

Evaluation of Computer-Use Agents (CUAs) is often limited to the final deliverables they create (at the end of hundreds of steps) and assessed with functional verifiers, as seen in OSWorld. However, such evaluation of end-state performance lacks transparency into \textit{how and why} agents fail in various tasks, obfuscating critical insight for subsequent improvement. For instance, agents that err during keyboard inputs would require a different mitigation strategy from those that fail to precisely provide click-based inputs on the graphical UI. We introduce OSWorld-Pro: a set of over 300 tasks containing over 2800 subgoals to enable the procedural evaluation of CUAs grounded in over 67,000 human annotations. We use robust human-aligned LLM-Judges to evaluate the fulfillment of OSWorld-Pro subgoals and thereby reveal the progress that models make throughout a series of sequentially dependent subgoals. Our findings reveal that OSWorld-Pro is challenging even for state-of-the-art LLMs, with top performers like Claude Opus 5 achieving only 75.7\% vs. 83.4\% on OSWorld. Furthermore, we identify critical process-focused failure modes of various models (e.g. subgoal-irrelevant actions and click-based mistakes) to provide insights to improve performance and efficiency of CUAs.

\end{abstract}

\definecolor{darkyellow}{rgb}{0.7, 0.7, 0.2}
\definecolor{darkgreen}{rgb}{0.463, 0.726, 0}

\begin{figure}[b]
    \centering
    \renewcommand{\arraystretch}{1.0}
    \small
    \vspace{-10pt}
    \includegraphics[width=\textwidth]{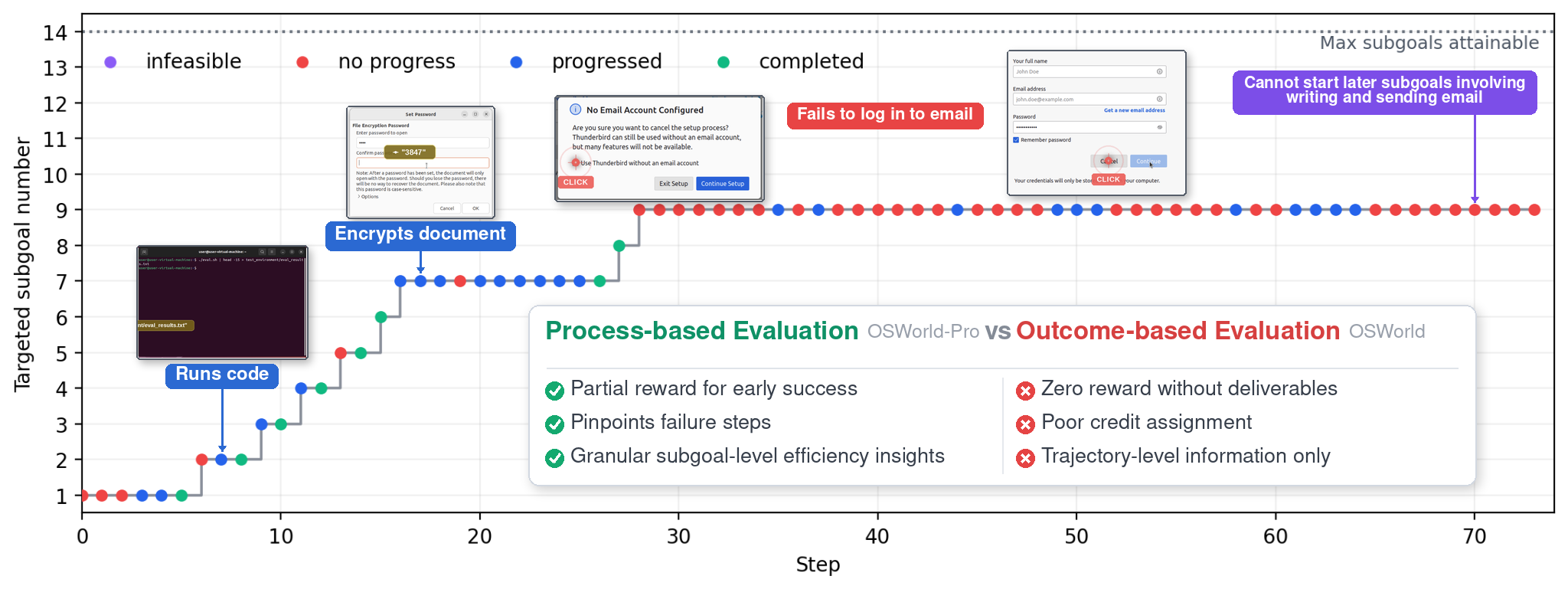}
\caption{OSWorld-Pro provides partial reward for early success without requiring final deliverables, pinpoints specific failures and offers granular information on progress through sequentially dependent subgoals. Such advantages of Process-based evaluation for Computer Use Agents (CUAs) complements the limitations of outcome-based evaluations such as OSWorld.}
    \label{fig:infographic}
\end{figure}

\section{Introduction}

Humans use computers to tackle long-horizon tasks involving many interdependent subgoals, tracking their progress through intermediate milestones rather than relying solely on final outcomes. For example, in a machine learning research project such as the one presented in this paper, researchers assess progress by monitoring data collection, iterating on computational experiments, and visualizing experimental results. In such situations, humans often do not rely on final outcomes alone (such as completed paper manuscripts) and instead utilize the status of various constituent subgoals to determine how these projects are progressing. Using the same vein of thought, we believe that process-based evaluations of Computer-Use Agents can complement existing benchmarks based on outcome-based evaluations (e.g. OSWorld). 

\begin{figure}[t]
    \centering
    \begin{tcolorbox}[
        width=\linewidth,
        colback=gray!3,
        colframe=darkgreen!65!black,
        boxrule=0.6pt,
        arc=2mm,
        left=2mm,
        right=2mm,
        top=1.5mm,
        bottom=1mm,
        title=\textbf{OSWorld-Pro Example (vs. OSWorld Example at bottom)},
        colbacktitle=darkgreen!55!white,
        coltitle=black,
        fonttitle=\small\bfseries,
        fontupper=\scriptsize
    ]

    \textbf{Category:} Coordination (between $\geq4$ different applications)
    
    \textbf{Goal:}
    Open Visual Studio Code from the sidebar. Create a new Python file
    named \texttt{calculate.py} in the Home directory that contains a
    function to calculate the factorial of a number and print the result.
    Run the script in the terminal with an input of 5. Then copy the output
    from the terminal and paste it into LibreOffice Writer. Format the
    pasted text as bold with font size 14, and save the document as
    \texttt{factorial\_result.docx} on the Desktop.

    \medskip
    \textbf{Subgoals:}
    \hfill
    \textbf{\textit{\color{gray}{[App used]}}}

    \begin{enumerate}
        \item Open Visual Studio Code from the sidebar.
              \hfill \textit{\color{gray}{[OS]}}

        \item Create \texttt{calculate.py} in the Home directory.
              \hfill \textit{\color{gray}{[VS Code]}}

        \item Write a function that calculates the factorial of a number
              and prints the result.
              \hfill \textit{\color{gray}{[VS Code]}}

        \item Run \texttt{calculate.py} in the terminal with an input of 5.
              \hfill \textit{\color{gray}{[Terminal]}}

        \item Copy the output from the terminal.
              \hfill \textit{\color{gray}{[Terminal]}}

        \item Paste the copied output into LibreOffice Writer.
              \hfill \textit{\color{gray}{[LibreOffice Writer]}}

        \item Format the pasted text as bold with font size 14.
              \hfill \textit{\color{gray}{[LibreOffice Writer]}}

        \item Save the document as \texttt{factorial\_result.docx}
              on the Desktop.
              \hfill \textit{\color{gray}{[LibreOffice Writer]}}
    \end{enumerate}

    \medskip

    \textbf{Steps:} (e.g. Step 1) \\
    \medskip
    \begin{minipage}[t]{0.1\linewidth}
    \end{minipage}
    \hfill
    \begin{minipage}[t]{0.47\linewidth}
        \vspace{0pt}
    
        \textbf{Action:}\\
        \texttt{pyautogui.click(0.019, 0.184)}
    
        \medskip
        \textbf{Reasoning:}\\
        ... The first step is to click on this icon to open Visual Studio Code. Let me click on the VS Code: icon to start the process.
    
        \medskip
        \textbf{Human Annotation:}\\
        \\
        Subgoal(s) targeted: \textbf{\{1\}}\\
        Subgoal(s) feasibility: \checkmark\\
        Subgoal(s) progression: \checkmark\\
        Subgoal(s) completion: \checkmark
    
    \end{minipage}
    \hfill
    \begin{minipage}[t]{0.49\linewidth}
        \vspace{0pt}
        \centering
    
        \includegraphics[
            width=\linewidth,
            keepaspectratio
        ]{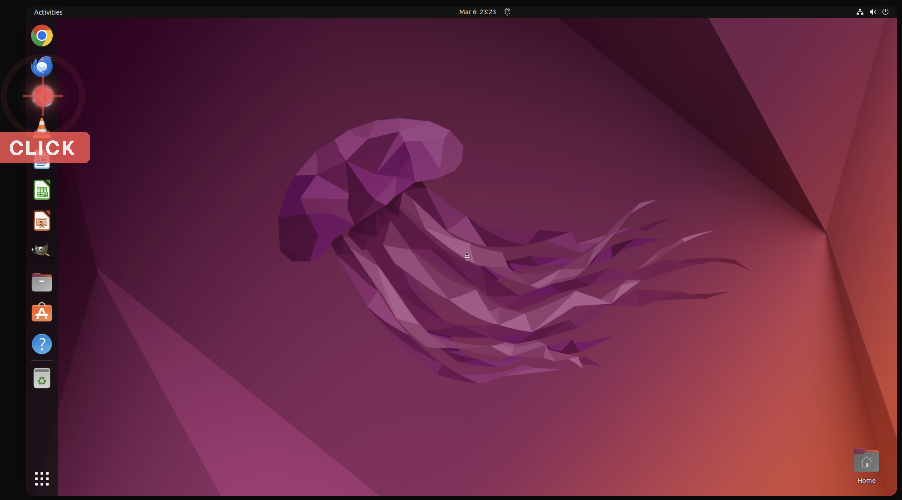}

    \end{minipage}
    \par\medskip
    \noindent\rule{\linewidth}{0.4pt}
    \smallskip
    \vspace{5pt}
    \textbf{OSWorld Example:} Make the line spacing of first two paragraph into double line spacing  \hfill \textit{\color{gray}{[LibreOffice Writer]}}
    \end{tcolorbox}

    \caption{OSWorld-Pro Example compared with OSWorld Example. More examples in
    \S~\ref{app:examples}.}
    \label{fig:front_example}
\end{figure}
\vspace{-7pt}

Outcome-based evaluations were utilized with great success on math capabilities in terms of GSM8K \citep{cobbe2021trainingverifierssolvemath} and AIME 25 \citep{livebench} and later in coding environments (using unit tests) such as LiveCodeBench \citep{jain2024livecodebenchholisticcontaminationfree} as well as scientific question answering in GPQA \citep{rein2023gpqagraduatelevelgoogleproofqa}, MMLU-Pro \citep{wang2024mmluprorobustchallengingmultitask},  and HLE \citep{phan2025humanitysexam}. Outcome-based evaluation was also applied on CUA capabilities in works such as the widely-adopted OSWorld \citep{xie2024osworld} and OSUniverse \citep{davydova2025osuniversebenchmarkmultimodalguinavigation}. These works construct functional verifiers against the final deliverables of the tasks and measure the success of agents based on whether the deliverables match  various aspects of reference files. Beyond desktop GUI agents, there have also been adjacent work relating to Android GUI \citep{rawles2025androidworlddynamicbenchmarkingenvironment} as well as tool calling ability with model context protocol \citep{jia2025osworldmcpbenchmarkingmcptool}. 

However, such outcome-based evaluation approaches for CUAs have some limitations. First, they are unable to discriminate between trajectories with different progression in tasks that have yet to create final deliverables. For instance, in Fig. \ref{fig:infographic}, final deliverables will not be available whether the agent fails at the first subgoal or the ninth subgoal and hence the agent will receive a zero for both trajectories even though the agent has progressed much further in the later trajectory. Such poor discernibility becomes more critical in long-horizon tasks (e.g. requiring hundreds of steps and therefore have a high likelihood of failure prior to final deliverable creation). 
Second, as CUAs become stronger, they will also be more capable in terms of reward hacking, which means that these agents can get to the correct final outcome through undesirable ways. The recent OpenAI security incident \citep{openairewardhacking} highlights how strong agents can literally hack third-party servers to obtain restricted information (i.e. answer keys) in order to do well on outcome-based evaluations. 
Finally, outcome-based evaluations do not provided fine-grained information on the contribution of individual steps within the agent trajectory, which can be useful to assess how efficiently agents  advance on long-horizon tasks. Process-based evaluation addresses these limitations by focusing not only on \textit{what} final outcomes CUAs arrive at but also \textit{how} they get there.

To design process-based evaluation for CUAs, we draw inspiration from PRM-800k \citep{lightman2023letsverifystepstep} and ProcessBench \citep{zheng-etal-2025-processbench}, which are process-based evaluations for math capabilities. Specifically, these benchmarks break down reasoning on solving math problems into distinct steps. Then, they seek to identify where errors first occur in the reasoning chain. CUA tasks tend to be much more open-ended compared to math tasks in PRM-800K and ProcessBench, as CUA tasks often have multiple approaches to reach the required goal. Therefore, we adapt ideas from works on rubrics \citep{gunjal2025rubricsrewardsreinforcementlearning, arora2025healthbenchevaluatinglargelanguage, wang2026profbench}. Specifically, we break down the overarching task goal into subgoals that can be individually assessed for completion. However, unlike rubrics, which are typically requirements that models can fulfill independently, subgoals in many CUA tasks are dependent on each other. This means that an earlier subgoal has to be completed before moving to the next subgoal.

To support process-based evaluation for CUAs, we present OSWorld-Pro: a benchmark with over 67, 000 step-level human annotations across more than 2800 progressive subgoals in over 300 long-horizon CUA tasks. The main features for OSWorld-Pro are:

1. \textbf{Long-Horizon}: Tasks have an average of 9.2 sequentially dependent subgoals per task, for which earlier subgoals have to be completed prior to attempt subsequent ones. This is in contrary to benchmarks like OSWorld where tasks mostly have a single subgoal (see example in Fig \ref{fig:front_example}) and ChainWorld \citep{siu2026chainworldcomposinglonghorizondesktop}, which chains up multiple loosely-connected OSWorld tasks and do not reflect the interdependent nature of subgoals in real-world long-horizon tasks.

2. \textbf{Challenging}: Tasks require an average of 3.45 unique apps to complete, which is substantially more compared to OSWorld at 1.34. In addition, we include rare apps such as Videos, Archive Manager and LibreOffice Draw and uncommon Linux and GUI distributions - beyond Ubuntu with GNOME - (e.g. AlmaLinux and MATE) not found in OSWorld to test generalization capabilities. Among top performing models, Claude Opus 5 only reaches 75.7\% vs. 83.4\% on OSWorld \citep{osworldwebsite}.

3. \textbf{Fine-grained}: Beyond providing an aggregate metric that shows how well an agent performs, OSWorld-Pro allows users to understand the type of actions that it commonly fails on, how efficient it is in completing subgoals over its trajectories and how it react when facing infeasible subgoals, as common in real-world tasks. For instance, Claude Opus 5 was shown to occasionally engage in subgoal-irrelevant actions (sometimes for over 50-steps) despite completing all subgoals.

\section{OSWorld-Pro Overview}

\begin{figure}
    \centering
    \renewcommand{\arraystretch}{1.0}
    \small
    \includegraphics[width=\textwidth]{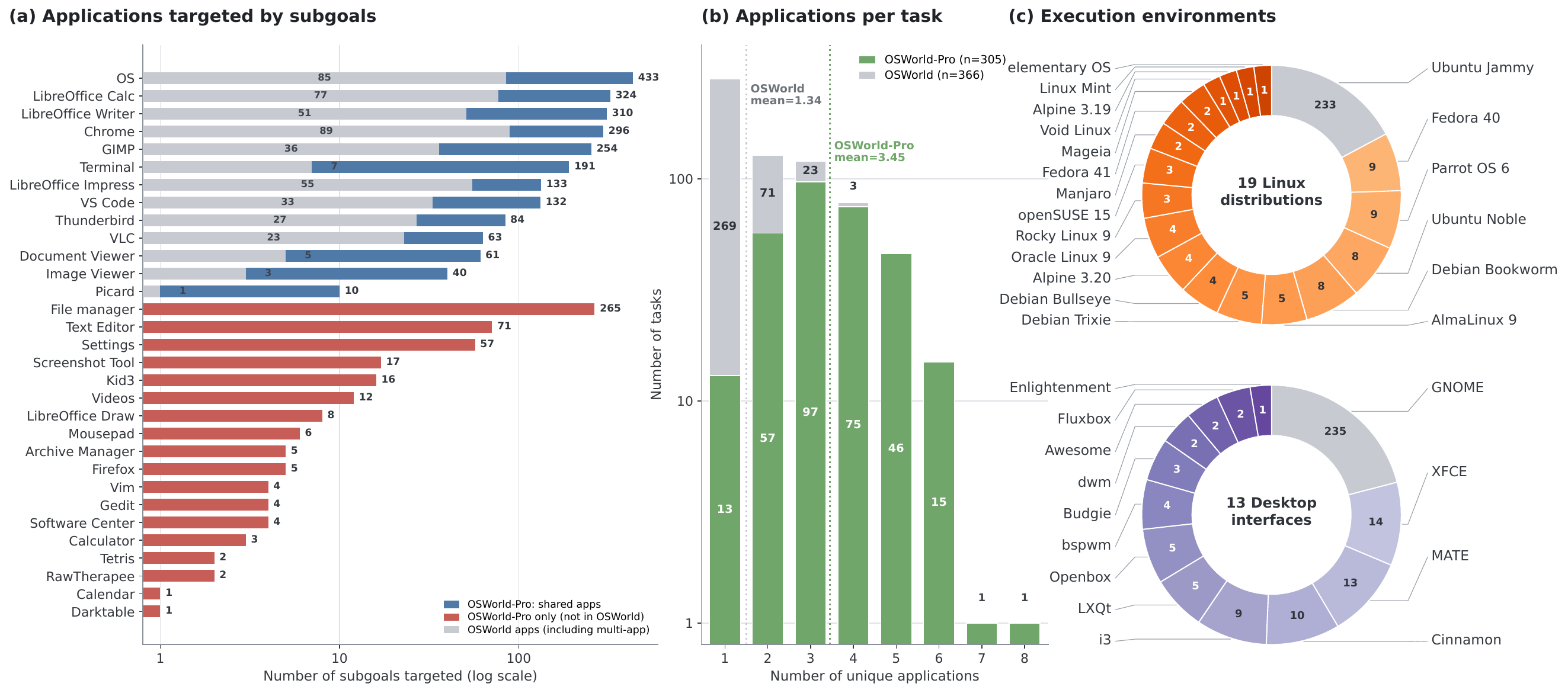}
\vspace{-20pt}
\caption{OSWorld-Pro Data Distribution: Compared with OSWorld, OSWorld-Pro covers a wider diversity of applications (31 vs 13), requires coordination between more applications in a single task (mean = 3.45 vs 1.34) and can show robustness of agents in more environments (19 Linux distributions + 13 Graphical interface vs Ubuntu Jammy with GNOME only) .}
    \vspace{-5pt}
    
    \label{fig:rubrics_distribution}
\end{figure}

OSWorld-Pro contains 67,264 human-annotated labels across 305 distinct tasks with 2814 progressive subgoals. These tasks span three categories: Diversity (117 tasks), covering less commonly benchmarked applications; Coordination (109 tasks), requiring coordination across $\geq 4$ applications; and Robustness (79 tasks), testing generalization across Linux distributions and graphical interfaces.  Specifically, samples contains agent trajectories on various tasks as well as step-level human annotations on how each step targets various subgoals, determine their feasibility and monitor their progression and final completion. While the sheer number of tasks (305) is comparable to the 369 tasks in OSWorld \citep{xie2024osworld} and 160 tasks in OSUniverse \citep{davydova2025osuniversebenchmarkmultimodalguinavigation}, each OSWorld-Pro task has 2 orders of magnitude more granular human-annotated verification signals (collected over $>5000$ person-hours) compared to alternatives. We show an example in Fig. \ref{fig:front_example}, a visualization of data distribution in Fig.~\ref{fig:rubrics_distribution}, and further descriptive statistics in \S \ref{app:descriptive_statistics}.

\vspace{-5pt}

\vspace{-5pt}
\section{Data Collection}\label{sec:data_collection}
\vspace{-5pt}
\paragraph{Annotator Recruitment} To ensure annotation quality, we select qualified annotators through screening, training and pairing each annotator with an experienced reviewer. Annotators with Bachelors' degrees or higher, as well as $\geq6$ months of experience working on Computer-Use Agent annotation are recruited and managed by our vendor. Prior to their inclusion into the project, we check that they pass tests on English capabilities, Linux proficiency (post mandatory training) and understanding of the annotation workflow (through two sample tasks reviewed against golden references). Each annotator works through the entire agent trajectory of a task and is supported by a reviewer (with substantial annotation experience) to iteratively review the annotator's work and to provide feedback for improvement where helpful. Across 305 tasks, 25 annotators and reviewers from 5 countries were involved. Further details on annotator recruitment in \S \ref{app:recruitment}.

\paragraph{Task Curation} 
We generate tasks using the approach described in ProCUA-SFT \citep{jung2026procuasfttechnicalreport} for human annotation. 
Specifically, for the Diversity category, we identify tasks containing applications that are underrepresented in existing computer-use benchmarks (e.g. Videos, Archive Manager and LibreOffice Draw).
For the Coordination category, we select tasks that require coordinated use of at least four applications. By comparison, tasks in OSWorld involve at most four applications. 
For the Robustness category, we generate additional tasks using the ProCUA-SFT approach\footnote{We use Kimi-K2.6, available in May 2026, instead of Kimi-K2.5.} in environments that differ from the default Ubuntu/GNOME setup. These environments include Linux distributions such as Fedora, Alpine, and AlmaLinux, and graphical interfaces such as bspwm, Xfce, and LXQt, as detailed in Fig.~\ref{fig:rubrics_distribution}. These tasks evaluate how well models generalize across Linux distributions and graphical interfaces.

\paragraph{Subgoal Decomposition} ProCUA-SFT tasks only contain an overarching task goal as well as the set of apps required for the task goal. We break down the tasks into atomic subgoals that can be independently assessed as well as a specific application required for each subgoal. Specifically, we prompt DeepSeek-V4-Pro \citep{deepseekai2026deepseekv4highlyefficientmilliontoken} using the prompt template in \S \ref{app:templates}.

\paragraph{Human Annotation} 
Our annotation workflow consists of task validation, annotator assignment, step-level labeling, and lastly independent and interactive review.
\textbf{(1) Task validation}
Prior to human annotations, our vendor inspects and removes tasks that have unclear or under-specified goals or have safety concerns. In addition, the subgoals and application required by each subgoal are also manually inspected and corrected where appropriate. In addition, we skip tasks where subgoals are not sequentially dependent on one another (i.e. only include tasks where earlier subgoals need to be completed before later subgoals). In doing so, we avoid overly simple tasks with multiple unrelated subgoals (e.g. open a video file, then open an unrelated text document) to focus on long-horizon tasks that are more challenging and realistic. 
\textbf{(2) Annotator assignment}
We assign tasks to technical or non-technical annotator pools based on the applications and skills involved. Tasks requiring coding knowledge, such as those involving coding in VS Code or the terminal, are assigned to technical annotators.
\textbf{(3) Step-level labeling}
Annotators label each step of pre-generated model trajectories to identify the subgoal(s)\footnote{This is almost always a single subgoal but $<$2\% of steps do effectively target two or more subgoals.} that a step targets. In some environments, we notice that the targeted subgoal is not feasible. For example, a subgoal may require selecting an option from the file menu that does not exist. Therefore, we also ask annotators to indicate whether the targeted subgoal(s) are feasible. Following OSWorld \citep{xie2024osworld}, we keep tasks with infeasible subgoals in order to understand what models would do in similar real-world tasks. If it is feasible, annotators also indicate if the current step makes progress on and completes the subgoal. 
\textbf{(4) Independent and interactive review}
Inspired by the annotation workflow in ProfBench \citep{wang2026profbench}, our workflow involves having a reviewer to iteratively provide feedback for the annotator to modify their annotations. However, because the task is time-consuming (annotators spend between 5 to 20 hours per task depending on the number of steps taken), we found it challenging for reviewers to provide comprehensive feedback, despite their strength in precisely pointing out inadequacies. Therefore, we modify our workflow based on the approach taken in HelpSteer3 \citep{wang-etal-2025-helpsteer3} such that the reviewer has to first independently annotate the task before providing feedback to the annotator with the annotation platform (Super Annotate) highlighting the differences in initial annotations. Annotators were not allowed to use LLMs during the annotation task, with extensive checks to ensure compliance (e.g. based on repetition patterns and annotation durations). Full annotation guidelines are in \S \ref{app:guidelines}.

\section{Can LLMs effectively judge like humans do?}\label{sec:llm_judge}

\paragraph{Task Formulation} Given the high cost of having humans evaluate agent trajectories, many works \citep{starace2025paperbenchevaluatingaisability, arora2025healthbenchevaluatinglargelanguage, wang2026profbench} have turned to using LLM-Judges to proxy human judgments. Computer Use Agent trajectories are particularly challenging to evaluate because they are path-dependent across long-horizons (e.g. across hundreds of steps) while evaluation of agent outputs (e.g. in OSWorld) is only state-dependent, without considering how the agent arrived at the final state. More precisely, the role of the LLM Judge is to identify which subgoal(s) are targeted by every step of the agent trajectory, whether that subgoal is feasible at the step and if so, whether the agent makes progress and/or completes the subgoal at that step. 

\vspace{-5pt}
\subsection{Evaluation}\label{sec:criterion_judge}
\vspace{-5pt}

\paragraph{Agreement with Human Annotations} To evaluate LLM-Judges, we use Macro-F1 based on the human-labeled ground-truth and the model-predicted label as used by ProfBench \citep{wang2026profbench} and PaperBench \citep{starace2025paperbenchevaluatingaisability}. Subgoal(s) targeted is first binarized into whether each subgoal is targeted by a particular step while the other fields (feasibility, progression and completion) are innately binary. Given the dependent nature of the fields, we consider values in fields iff the pre-requisite field was correct. This means we consider feasibility iff the correct set of subgoals were identified, progression iff feasibility was correct and completion iff progression was correct.

\paragraph{Inference Setup}

We ran early experiments with GPT-5.6-Luna (with default medium reasoning), a lightweight, affordable and feasible at high concurrency. The complexity of judgment task (with hundreds of step over tens of subgoals) required extensive explorations in LLM-judge design, which we discuss in \S\ref{app:llm_judge}.
Our final design involves evaluating an entire trajectory in a single API request. Because this requires sending a payload of up to hundreds of screenshots in a single request (requiring 500 MBs), we only found success in using OpenAI GPT-5.6 models while others (e.g. Claude, Gemini and Open Models) raised various errors relating to payload size, image quantity or context window. Cost estimation method is detailed in \S \ref{app:inference_setup}.

\paragraph{Performance at Different Levels}
In addition to Macro-F1 at the step level, we want to understand how well the LLM-Judge can predict human annotations at three complementary levels. First at the step level, we calculate a simple mean across macro-F1 across subgoal \{targeted, feasibility, progression, completion\}. Next at the subgoal level, we tabulate the proportion of subgoals that have their final completion status across all steps in a trajectory predicted correctly. Subgoals are considered completed iff at least one step has the targeted subgoal labeled as completed. Finally, at the task level, we calculate an overage task performance based on the percentage of subgoals completed. We calculate a mean absolute error (MAE) between the human-annotated and model-predicted performance and take the average across all tasks. Finally, we calculate 1-MAE to make the metric higher as better and improve readability.

\paragraph{Reliability of Human Annotations} To understand how reliable the Annotators' labels are, we compare them to the Reviewer's initial annotations prior to see annotations from the Annotator and providing feedback to the Annotator. This provide a measure of how much two independent annotators would agree. We find excellent agreement (Cohen's $\kappa$ $>$ 0.8) across all aspects with 0.987 for subgoal targeted, 0.847 for feasibility, 0.869 for progression and 0.940 for completion.

\vspace{-5pt}
\subsection{Results}\label{sec:judge_results}
\vspace{-15pt}
\definecolor{lightyellow}{RGB}{255,255,200}

\begin{table}[h!]
\centering
\caption[Judge Models]{Evaluation of LLM-Judges. Higher is better for Macro-F1 and Performance at different levels while lower is better for Tokens.}
\begin{adjustbox}{max width=\columnwidth, scale=1
}
\begin{tabular}{l|cccc|ccc|ccc}
\toprule
& \multicolumn{4}{c|}{\textbf{Step Agreement w. Human Labels (Macro-F1) $\uparrow$ }} & \multicolumn{3}{c}{\textbf{Performance at Different Levels $\uparrow$}} & \multicolumn{3}{c}{\textbf{Tokens $\downarrow$} }\\

\textit{Model} & Target & Feasible & Progress & Complete & Step (Mean) & Subgoal & Task (1-MAE) &  In/Task & Out/Task & Total \$\\
\midrule
\textit{\textbf{Human Performance}} &  98.4 & 91.6 & 94.1 & 97.6 & 95.4 & 95.6 & 96.0 & - & - & - \\
\midrule
\textbf{\textit{LLM Judge}} \\
\midrule
OpenAI/GPT-5.6-Sol \\
- max & 97.0 & 61.9 & 74.7 & 94.3 & 82.0 & 94.1 & 93.0 & 144566 & 12004 & 249.60\\
- xhigh & 97.2 & 65.3 & 71.6 & 93.8 & 82.0 & 93.9 & 93.0 & 144566 & 7072 & 219.51\\
- high & 97.0 & 65.8 & 69.8 & 93.3 & 81.5 & 93.7 & 92.6 & 144566 & 4900 & 206.26\\
- medium & 96.8 & 64.1 & 68.1 & 91.3 & 80.1 & 92.9 & 92.5 & 144566 & 3631 & 198.52\\
- low & 95.8 & 65.9 & 65.2 & 86.8 & 78.4 & 89.1 & 90.5 & 144566 & 2706 & 192.88\\
- none & 94.5 & 67.5 & 69.5 & 67.8 & 74.8 & 83.8 & 86.5 & 144566 & 2069 & 188.99\\

OpenAI/GPT-5.6-Terra  \\
- max & 97.6 & 62.0 & 69.2 & 94.8 & 80.9 & 93.9 & 92.4 & 144566 & 18266 & 155.04\\
- xhigh & 97.2 & 64.5 & 63.9 & 91.6 & 79.3 & 92.6 & 90.9 & 144566 & 7192 & 114.51\\
- high & 95.9 & 63.8 & 62.3 & 86.2 & 77.1 & 92.2 & 91.3 & 144566 & 4168 & 103.44\\
- medium & 94.9 & 67.4 & 59.6 & 82.9 & 76.2 & 89.6 & 90.7 & 144566 & 2925 & 98.89\\
- low & 94.9 & 66.0 & 61.3 & 80.0 & 75.6 & 89.1 & 89.8 & 144566 & 2609 & 97.74\\
- none & 93.3 & 65.2 & 67.6 & 82.8 & 77.2 & 91.5 & 91.3 & 144566 & 1961 & 95.36\\

OpenAI/GPT-5.6-Luna \\
- max & 97.7 & 58.9 & 52.8 & 93.3 & 75.7 & 91.2 & 89.6 & 144566 & 18111 & 15.45\\
- xhigh & 97.1 & 60.4 & 51.0 & 89.6 & 74.5 & 87.3 & 87.4 & 144566 & 11736 & 13.11\\
- high & 96.5 & 58.8 & 49.8 & 86.5 & 72.9 & 84.0 & 84.9 & 144566 & 6713 & 11.28\\
- medium & 95.6 & 57.3 & 47.6 & 84.6 & 71.3 & 85.1 & 87.5 & 144566 & 2942 & 9.90\\
- low & 94.9 & 54.0 & 46.4 & 87.0 & 70.6 & 88.5 & 89.0 & 144566 & 2309 & 9.66\\
- none & 90.4 & 58.9 & 53.6 & 81.2 & 71.0 & 84.3 & 86.7 & 144566 & 1932 & 9.53\\

\bottomrule
\end{tabular}
\end{adjustbox}
\label{tab:llm_judge_evaluation}
\end{table}
\vspace{-10pt}

\paragraph{Which aspects are LLM Judges lacking compared to humans?} The top performing model GPT-5.6-Sol with max reasoning effort approaches Human Performance on identifying targeted subgoals (97.0 vs 98.4\%) and comes close on predicting subgoal completion at the step, subgoal and task levels (within 3.3\% absolute different from humans) as shown in Tab. \ref{tab:llm_judge_evaluation}. However, it substantially lags behind on predicting subgoal feasibility (61.9 vs. 91.6\%) and progress (74.7 vs. 94.1\%), which also have lower agreement rates between independent human annotators (Cohen's $\kappa$=0.847-0.869 vs 0.940-0.987). We suspect it is because these fields are more subjective as it might not always be clear whether a task is infeasible or a practical method to achieve a goal has not yet been found. For example, when locating an option under a menu with various unexpanded sub-menus, a feasible subgoal might appear infeasible until the menu option is discovered (or vice-versa). Similarly, whether a step meaningfully helps to progress toward a subgoal might at times be ambiguous. For instance, if a step modifies a piece of code to add a new functionality but results in a bug, a judge can reasonably justify both progress=No or progress=Yes depending on what the focus is on. 

\paragraph{Does Model Size Matter?}

Within the GPT-5.6 model family, larger models generally do better than smaller models but the improvement plateaus. For instance, the step level performance (Mean of 4 Macro-F1s) in Tab. \ref{tab:llm_judge_evaluation} substantially increase from 75.7 (Luna) to 80.9 (Terra) and then only slightly increases to 82.0 (Sol) while at Max reasoning. This rate of increase is roughly inline with gains in total cost, which first increases 10x and then rises only 1.6x. When constrained on API cost, increasing the reasoning effort on a smaller model often beats a lower reasoning effort on a larger model at a lower cost. For instance, GPT-5.6-Terra Max slightly outperforms GPT-5.6-Sol Medium at 22\% lower price while GPT-5.6-Luna Max generally matches GPT-5.6-Terra Low at one-sixth cost.

\paragraph{How much should a model think?}

Models generally perform better at higher reasoning efforts across most metrics. However, there are two major exceptions to the rule. Models sometimes perform better with no reasoning compared to low reasoning. This might be because given models insufficient thinking tokens might artificially cut short its explicit thinking process, while non-reasoning model are unaffected in their implicit, latent reasoning through hidden layers \citep{li2025implicitreasoninglargelanguage}. Subgoal Feasibility performance generally also lowers with higher reasoning effort (on Terra and Sol). Such a correlation is also observed in terms of the likelihood for models to predict feasibility=No. The proportion of feasibility=No decreases from 9.7\% at GPT-5.6-Sol None to 5.1\% at GPT-5.6-Sol Max while human-annotated ground-truth is at 15.3\%. This suggests that increased reasoning effort results in models over-estimating the feasibility of tasks and hence resulting in lower macro-F1.

\section{Benchmarking Models on OSWorld-Pro}\label{sec:report_generator}
\vspace{-15pt}

\begin{table}[h!]
\centering
\caption[Response Generation Systems]{OSWorld-Pro Performance across select Closed-Source and Open-Weight models.}
\begin{adjustbox}{max width=\columnwidth, scale=1
}
\begin{tabular}{l|cccc|ccccc|ccc}
\toprule
& \multicolumn{4}{c|}{\textbf{\% of Tasks with All Subgoals Completed $\uparrow$}} & \multicolumn{3}{|c}{\textbf{Task-Level Efficiency $\downarrow$} }\\ 

\textit{Model} & Overall & Diversity & Coordination & Robustness  &  Steps & Output Tokens & Token \$\\

\midrule

Closed-Source\\
\midrule
Claude Opus 5 Max & 75.7 & 81.2 & 70.6 & 74.7  & 76.0 & 23492.6 & 3.36 \\

Claude Opus 4.8 Max & 77.7 & 78.6 & 83.5 & 68.4  & 67.1 & 67069.6 &  5.60 \\ 
Claude Opus 4.7 Max & 76.7 & 78.6 & 80.7 & 68.4 &  98.8 & 30886.2 & 4.95 \\ 
Claude Sonnet 5 Max & 76.4 & 82.1 & 81.7 & 60.8 &  108.7 & 40804.6 &  2.47 \\ 
GPT-5.6-Sol Max & 74.8 & 76.1 & 70.6 & 78.5 &  53.3 & 39595.2 & 9.56 \\ 
GPT-5.6-Terra Max & 70.8 & 74.4 & 71.6 & 64.6 &  54.9 & 43777.5 & 4.98\\ 
GPT-5.6-Luna Max & 74.4 & 74.4 & 74.3 & 74.7 &  60.2 & 41377.1 & 0.51 \\ 
Gemini-3.8-Flash High & 59.0 & 65.8 & 59.6 & 48.1 &  71.7 & 18966.8 & 2.81 \\ 
\midrule
Open-Weight\\
\midrule
Kimi K3 Max (2.8T) &39.3 & 46.2 & 43.1 & 24.1 &  42.2 & 125870.7 & 5.51 \\ 
Minimax M3 Xhigh (428B) & 28.9 & 40.2 & 30.3 & 10.1  & 201.3 & 36952.1 & 1.04 \\ 
Qwen 3.8 Flash Next Xhigh (125B) & 55.1 & 66.7 & 58.7 & 32.9 &  75.0 & 26940.4 & 0.20 \\ 
Qwen 3.5 122B Xhigh & 16.7 & 32.5 & 11.0 & 1.3 & 127.2 & 42316.4 & 0.68 \\ 
Qwen 3.8 27B Xhigh & 32.1 & 43.6 & 33.9 & 12.7 & 48.5 & 15177.7 & 0.15\\ 
Qwen 3.6 27B Xhigh & 10.2 & 20.5 & 6.4 & 0.0 & 31.5 & 7075.7 & 0.16 \\ 
\bottomrule
\end{tabular}
\end{adjustbox}

\label{tab:response_generation_evaluation}
\end{table}

\paragraph{Task Formulation} Following \citet{xie2024osworld} and \citet{jung2026procuasfttechnicalreport}, we formalize the task as given a goal alongside an computer-use Linux environment with Graphical UI (that it can "see" via screenshots), the agent (VLM with harness defined in OSWorld) should generate a trajectory that addresses the goal. Subsequently, we use the best-performing GPT-5.6-Sol Max judge from \S \ref{sec:llm_judge} to evaluate whether steps fulfill various subgoals. Finally, we report the percentage of tasks across each category that complete all subgoals deemed feasible by human annotators. We believe that GPT-5.6-Sol Max judge is an adequate proxy of human judgments as it matches human judgment in 94.1\%, closely trailing independent reviewers at 95.6\% from Tab. \ref{tab:llm_judge_evaluation}. In addition, GPT-5.6-Sol Max scores itself lower than 4 other models, alleviating our initial concerns over potential self-preference bias. Our evaluation setup largely follows OSWorld GitHub \citep{osworldgithub} and evaluates only vision-language models with publicly available harnesses there (detailed in \S \ref{app:inference_setup}).

\paragraph{OSWorld-Pro is Challenging} Overall, Claude Opus 4.8 with Max effort achieves the best performance in Tab. \ref{tab:response_generation_evaluation} at 77.7\% Overall, which means that OSWorld-Pro is substantially more challenging than OSWorld where the top Opus model scores 83.4\% \citep{osworldwebsite}. OSWorld-Pro is particularly challenging for open-weight models with the top model only achieving 55.1\% whereas open-weight models scores $>$80\% on OSWorld. For a matched model, Minimax M3 scores only 28.9\% on OSWorld-Pro but 75.2\% on OSWorld. This indicates that OSWorld-Pro can be a good target for open-weight models to hill-climb against, without being saturated or overly-difficult such that improvements are hard to achieve.

Among open-weight models, scores are generally highest on the Diversity category (least challenging), followed by Coordination and finally Robustness (most challenging). This means that open-weight models generalize best to rare applications, moderately to longer-horizon workflows requiring coordination between $\geq 4$ apps and worst on different Linux distributions and graphical interfaces. One possible explanation for the poor performance relating to generalizations across different Linux and graphical UI environment is the general lack of such data within training environments, due to the limited commercial advantage of improving on them (given how esoteric they are in real-world work settings). Therefore, models are forced to generalize out-of-distribution from  Ubuntu/GNOME environments that are more common in training data \citep{wang2025opencua, jung2026procuasfttechnicalreport}. 

\paragraph{Does Parameter-Scaling Work?} To understand the role that model-size plays on OSWorld-Pro performance, we conduct some analysis across both closed-source and open-weight models. Closed-source models in general perform better compared to open-weights one. As the parameter counts for closed models are not publicly reported, we use the observation that within the same model family, more expensive models by API pricing are likely to be larger (e.g. GPT-5.6 Sol $>$ Terra $>$ Luna, Opus 5 $>$ Sonnet 5). We find no obvious evidence that larger parameter count alone translates to better performance on OSWorld-Pro given that Sonnet 5 does better than Opus 5 (but worse than Opus 4.7 and 4.8) and GPT-5.6 Luna does better than Terra but worse than Sol. 

However, we observe that smaller model typically have a larger step count (e.g. Luna with 60.2 steps vs. Sol at 54.9 steps; Sonnet 5 at 108.7 steps vs. Opus 5 at 76.0 steps). This suggests that smaller models can use a greater number of steps to partially compensate for model capacity proxied via parameter count. On open-weight models, parameter count could potentially contribute some improvement as Qwen 3.8 Flash Next (125B) does better than Qwen 3.8B 27B at 55.1\% vs. 32.1\% from the same model family but it could be confounded by the difference in model architecture (Dense vs MoE). Furthermore, models with similar parameter count (e.g. Qwen 3.8 27B vs. Qwen 3.6 27B) have drastic different performance at 32.1\% vs. 10.2\% as with Qwen 3.8 Flash Next 125B and Qwen 3.5 122B (55.1\% vs. 16.7\%), suggesting that training recipes could influence performance more compared to model size alone.

\paragraph{Task-Level Efficiency} is critical in real-world tasks (beyond task completion) as it influences user experiences in terms of latency and additional cost. There are three complementary perspectives for users who care about different aspects: steps, output tokens and token cost. Latency-sensitive users should focus on a combination on the number of steps as well as the output-tokens while token costs should be the priority for cost-sensitive users. One observation is that some models have low step count but high output tokens (e.g. Kimi K3 with 125 thousand tokens in only 42.2 steps). A possible explanation is that the pyautogui library that evaluation depends on supports multiple actions per step and hence models like Kimi K3 does fewer steps overall but seeks to perform more in each step, which requires more response tokens (of which, many are used for thinking). Another observation is that cost per task can differ by 20x across models with similar performance (\$0.51 for GPT-5.6 Luna vs. \$9.56 for GPT-5.6-Sol), suggesting that good performance can be balanced with affordability.

\begin{table}[h!]
\centering
\caption[Error Analysis]{Analysis of Subgoal Progression Likelihood by Action Category as well as Efficiency}
\begin{adjustbox}{max width=\columnwidth, scale=1
}
\begin{tabular}{l|ccccccc|c|ccc|cc|ccc|ccc}
\toprule
& \multicolumn{7}{|c|}{\textbf{Subgoal Progression Likelihood $\uparrow$}}
& \textbf{\% Subgoals $\uparrow$} 
& \multicolumn{3}{c|}{\textbf{Subgoal Efficiency $\downarrow$}} \\
\textit{Model}
& All
& Click
& Drag
& Keyboard
& Scroll
& Execution 
& Others & Completed & Successful & Failure & Infeasible\\
&&&+Move&&&Control&& (Feasible Only)& Effort & Persistence & Persistence\\
\midrule
Claude Opus 5 Max & 86.5 & 89.0 & 73.9 & 92.9 & 92.9 & 78.3 & 83.9 & 85.6 & 7.2 & 10.5 & 9.5 \\

Claude Opus 4.8 Max & 89.6 & 92.5 & 74.3 & 92.6 & 93.8 & 84.5 & 85.4 & 86.7 & 6.7 & 12.1 & 12.1 \\
Claude Opus 4.7 Max & 80.7 & 86.3 & 41.0 & 89.2 & 84.1 & 69.9 & 73.9 & 91.0 & 8.7 & 18.9 & 32.2 \\
Claude Sonnet 5 Max & 85.5 & 88.0 & 44.9 & 93.0 & 91.1 & 73.4 & 81.8 & 89.2 & 9.9 & 28.7 & 39.8 \\
GPT-5.6-Sol Max  & 78.9 & 86.9 & 66.5 & 73.7 & 74.6 & - & 71.7 & 93.1 & 4.9 & 8.2 & 17.8 \\
GPT-5.6-Terra Max  & 79.3 & 86.9 & 61.2 & 74.4 & 63.1 & - & 73.8 & 92.8 & 5.3 & 9.1 & 12.1 \\ 
GPT-5.6-Luna Max  &77.7 & 84.7 & 57.3 & 73.1 & 74.2 & - & 79.2 & 93.2 & 5.6 & 10.7 & 26.8 \\
Gemini 3.8 Flash High & 74.4 & 83.1 & 76.8 & 73.4 & - & 82.1 & 64.4 & 67.9 & 8.1 & 18.6 & 20.5 \\
\midrule
Open-Weight\\
\midrule
Kimi K3 Max (2.8T) & 54.5 & 41.9 & 38.7 & 54.8 & 48.1 & 65.3 & 58.4 & 58.0 & 5.1 & 12.5 & 20.2 \\
Minimax M3 Xhigh (428B) & 49.7 & 39.0 & 24.2 & 78.2 & 66.6 & 40.4 & 57.2 & 52.5 & 21.2 & 64.3 & 80.7 \\
Qwen 3.8 Flash Next Xhigh (125B) & 73.4 & 72.4 & 53.9 & 76.6 & 78.6 & 70.7 & 50.2 & 78.2 & 7.3 & 19.9 & 22.5 \\
Qwen 3.5 122B Xhigh & 32.3 & 29.6 & 11.3 & 41.4 & 34.7 & 35.4 & 5.9 & 42.7 & 6.4 & 62.8 & 24.3 \\
Qwen 3.8 27B Xhigh & 80.0 & 82.7 & 51.6 & 79.3 & 70.2 & 84.8 & 52.2 & 54.9 & 6.3 & 12.3 & 32.7 \\
Qwen 3.6 27B Xhigh &65.4 & 65.9 & 55.6 & 63.1 & 58.8 & 73.7 & 58.3 & 29.0 & 5.5 & 9.9 & 12.5 \\
\bottomrule
\end{tabular}
\end{adjustbox}

\label{tab:ablation}
\end{table}
\vspace{-6pt}

\section{Analysis: What kind of actions and subgoals trip up models?}

 Process-based evaluations (like OSWorld-Pro) have unique advantages in revealing insights on efficiency and failures over an agent trajectory that outcome-based evaluation like OSWorld cannot.
 
\paragraph{Action-Level Failure Analysis} 
To identify which action types are most prone to failure for each model, we measure action-level progression likelihoods across agent trajectories.
Specifically, we calculate the proportion of steps that make progress toward the subgoals. 
A step is considered if it targets at least one subgoal and the targeted subgoals are feasible. We exclude action categories that were observed fewer than 5 times to reduce noise from limited observations. Higher scores are better.

\begin{figure}[h]
    \centering
    \clipbox{0pt {.5\height} 0pt 0pt}{%
        \includegraphics[width=\textwidth]{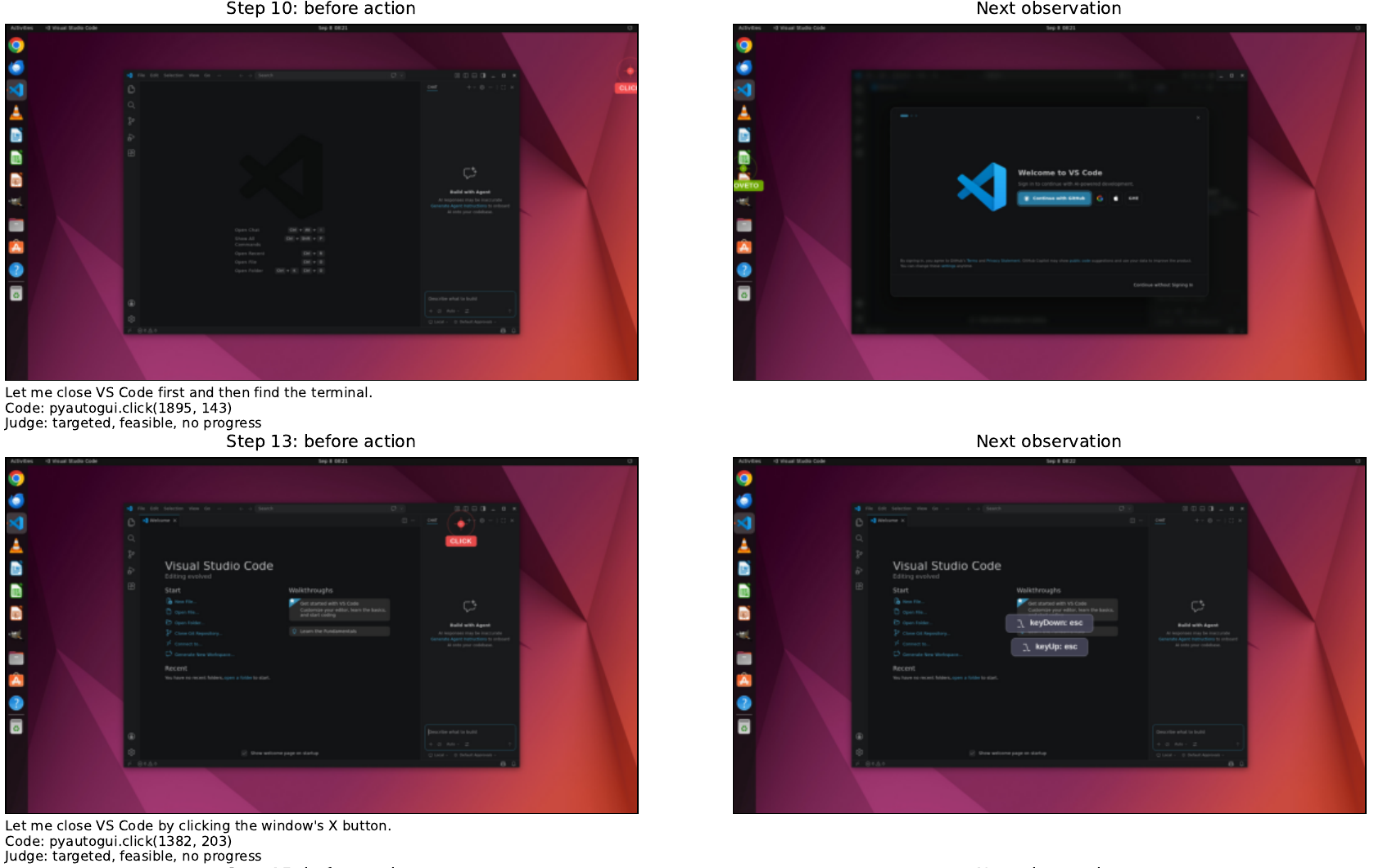}
    }
    \caption{Minimax M3 fails to click on the correct coordinates in order to perform the desired action.}
    \label{fig:minimax_click_failure}
\end{figure}
\vspace{-7pt}

Most models perform well on click operations, with the exception of Minimax M3 and Kimi K3. A closer look at Minimax M3 trajectories show frequent failures in basic operations such as clicking on the correct coordinates to perform an action (e.g. wanting to close a window but not clicking on the x button as shown in Fig. \ref{fig:minimax_click_failure}) while stronger models such as Claude and GPT-5.6 rarely commit such errors. This is reflected in the difference in progression likelihood for click operations on Tab. \ref{tab:ablation} for these models (39.0-41.9\%) vs. stronger models (72.4-92.5\%). 
In addition, Claude Opus 4.8 made substantial improvements on Click (86.3 to 92.5\%) in addition to Drag+Move (41.0 to 74.3\%) category over its predecessor Opus 4.7. This indicates the likelihood for purposeful training relating to pointer behavior for Opus 4.8.
Claude models are also excellent on Keyboard and Scroll type actions, beating out all other models by a healthy margin. For instance, Claude models score $\geq$ 89.2\% on Keyboard and $\geq$ 84.1\% on Scroll behavior while no other model reaches 80\% on either category. Among tested open-weight models, Qwen 3.8 models stand out despite being much smaller (27B to 125B) compared to other models (428B to 2.8T) as well as their own prior generations of similar sizes (Qwen 3.5 and 3.6), suggesting purposeful training on Qwen 3.8 to do well on GUI manipulation.

\begin{figure}[b]
    \centering
    \renewcommand{\arraystretch}{1.0}
    \small
    \includegraphics[width=\textwidth]{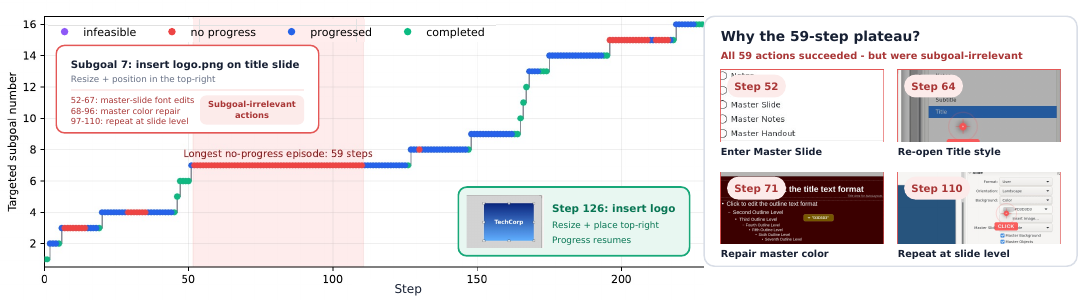}
    \caption{OSWorld-Pro reveals inefficiency of strong models such as Claude Opus 5 Max beyond what outcome-based evaluation alone (e.g. in OSWorld) can show. Opus 5 Max was stuck on multiple subgoals without any progress including once for 59 steps as it performed goal-irrelevant actions.}
    
    \label{fig:opus_inefficiency}
\end{figure}
\paragraph{Subgoal-Level Efficiency} Inspections of OSWorld-Pro trajectories reveal that models can show extremely inefficient behavior despite completing all subgoals, with an example from Claude Opus 5 Max in Fig. \ref{fig:opus_inefficiency}. To quantify such insights, we consider the average number of steps that the model spends on subgoals of varying status - completed (Successful Effort), feasible but not completed (Failure Persistence) and attempted but not feasible (Infeasible Persistence).  Successful Effort represents how directly models complete tasks without detours or unnecessary verification. At this stage, GPT-5.6-Sol and Kimi K3 are efficient at 4.9 and 5.1 steps/subgoal compared to Sonnet 5 and Minimax M3 at 9.9 and 21.2 steps respectively. Failure Persistence represents how hard models try when they fail to eventually complete a feasible subgoal, which broadly correlates well with model efficiency on Successful Effort. Infeasible Persistence represents how quickly models recognize infeasible tasks and give up. Some models like Opus 5 (9.5 steps/subgoal) and GPT-5.6-Terra (12.1) recognize infeasibility rapidly while others persist for more steps (e.g. Sonnet 5 at 39.8). 

\section{Conclusion}

We present OSWorld-Pro, the first process-based evaluation benchmark for Computer-Use Agents (CUAs) to complement outcome-based evaluations such as OSWorld. Based on over 67,000 human-annotations, 
OSWorld-Pro is a challenging benchmark - especially for open-weight models - that  reveals insights into failure modes that have previously eluded outcome-based evaluations (e.g. subgoal-irrelevant behavior and click-based operations). 
We believe OSWorld-Pro is a critical step to improving CUA evaluation that also comes with potential subsequent applications to guide the performance and efficiency improvement of CUAs (through harness optimization or process reward signals in  reinforcement learning), which we leave as future work.

\newpage

\subsection*{AI use statement}

In this work, we used generative AI tools for the following tasks:

\begin{enumerate}
    \item Generate synthetic data sets
    \item Implement methods
    \item Clean and reformat dataset
    \item Support qualitative and thematic data analysis
\end{enumerate}

We have not used generative AI tools for the following tasks:

\begin{enumerate}
    \item  Help develop theoretical models or conceptual frameworks
    \item Propose or refine hypotheses
    \item Interpret results
    \item Design or provide feedback on research  methodology or experiments
\end{enumerate}

The remaining disclosure tasks are not applicable to this work:

\begin{enumerate}
    \item Formulate mathematical claims
    \item Provide critical ingredients for proving mathematical claims
    \item Assist in the writing of proofs
    \item Assist with translation
\end{enumerate}

Additionally, we used generative AI tools for:

\begin{enumerate}
    \item Create or modify scientific figures or images
    \item Create or edit software code
\end{enumerate}

We have reviewed all AI-assisted work. 

\begin{enumerate}
    \item LLM-generated code was verified and tested for correctness by 2 authors
    \item Data visualizations were checked by authors against the supplied data to ensure data integrity
    \item Human annotators verified synthetic datasets generated, cleaned and reformatted with AI
\end{enumerate}

We take responsibility for the final content of this work,
including text, claims or artifacts produced with the aid of generative AI.

\section*{Ethics Statement}

All data collection carried out on this project was performed by our vendor, following internal reviews on  ethical and legal standards prior to the start of the project. All annotators engaged for this project were provided with transparent pay rates before work begins, timely payment on a fixed schedule as well as reasonable working hours and break guidance. If needed, annotators have access to confidential escalation paths for concerns as well as the removal of work content that may create undue risk without additional safeguards. All annotators were paid in accordance to applicable local labor laws as well as internal standards for worker protection and fair compensation.

\section*{Reproducibility statement}

Procedures for data collection has been extensively documented in \S \ref{sec:data_collection}, \ref{app:descriptive_statistics} and \ref{app:recruitment}. Evaluation details are in \S \ref{sec:criterion_judge}, \ref{sec:report_generator} and  \ref{app:templates}.

\bibliography{iclr2027_conference}
\bibliographystyle{iclr2027_conference}

\newpage

\appendix

\section{Example Data}\label{app:examples}

\begin{figure}[t]
    \centering
    \begin{tcolorbox}[
        width=\linewidth,
        colback=gray!3,
        colframe=blue!20!white,
        boxrule=0.6pt,
        arc=2mm,
        left=2mm,
        right=2mm,
        top=1.5mm,
        bottom=1mm,
        title=\textbf{OSWorld-Pro Diversity Example},
        colbacktitle=blue!20!white,
        coltitle=black,
        fonttitle=\small\bfseries,
        fontupper=\scriptsize
    ]

    \textbf{Rare App:} Kid3 - audio tagger for music metadata (not seen in OSWorld)
    
    \textbf{Goal:}
    Install the Kid3 audio tagging application from the KDE website and use it to edit the metadata for Zhou Xuan - Nights in Shanghai.mp3 in the Music folder, adding the album name and release year.

    \medskip
    \textbf{Subgoals:}
    \hfill
    \textbf{\textit{\color{gray}{[App used]}}}

    \begin{enumerate}
        \item Download the Kid3 application from the KDE website.
              \hfill \textit{\color{gray}{[Chrome]}}

        \item Install the Kid3 application.
              \hfill \textit{\color{gray}{[Terminal]}}

        \item Open Zhou Xuan - Nights in Shanghai.mp3 from the Music folder in Kid3.
              \hfill \textit{\color{gray}{[Kid3]}}

        \item Add the album name to the metadata of the MP3 file.
              \hfill \textit{\color{gray}{[Kid3]}}

        \item Add the release year to the metadata of the MP3 file.
              \hfill \textit{\color{gray}{[Kid3]}}

    \end{enumerate}

    \medskip

    \textbf{Steps:} (e.g. Step 1) \\
    \medskip
    \begin{minipage}[t]{0.1\linewidth}
    \end{minipage}
    \hfill
    \begin{minipage}[t]{0.47\linewidth}
        \vspace{0pt}
    
        \textbf{Action:}\\
        \texttt{pyautogui.click(0.500,0.556)}
    
        \medskip
        \textbf{Reasoning:}\\
        ... I can see the Kid3 website is already open in one of the tabs. Let me click on that tab to see the full page and find download instructions
    
        \medskip
        \textbf{Human Annotation:}\\
        \\
        Subgoal(s) targeted: \textbf{\{1\}}\\
        Subgoal(s) feasibility: \checkmark\\
        Subgoal(s) progression: \checkmark\\
        Subgoal(s) completion: \ding{55}
    
    \end{minipage}
    \hfill
    \begin{minipage}[t]{0.49\linewidth}
        \vspace{0pt}
        \centering
    
        \includegraphics[
            width=\linewidth,
            keepaspectratio
        ]{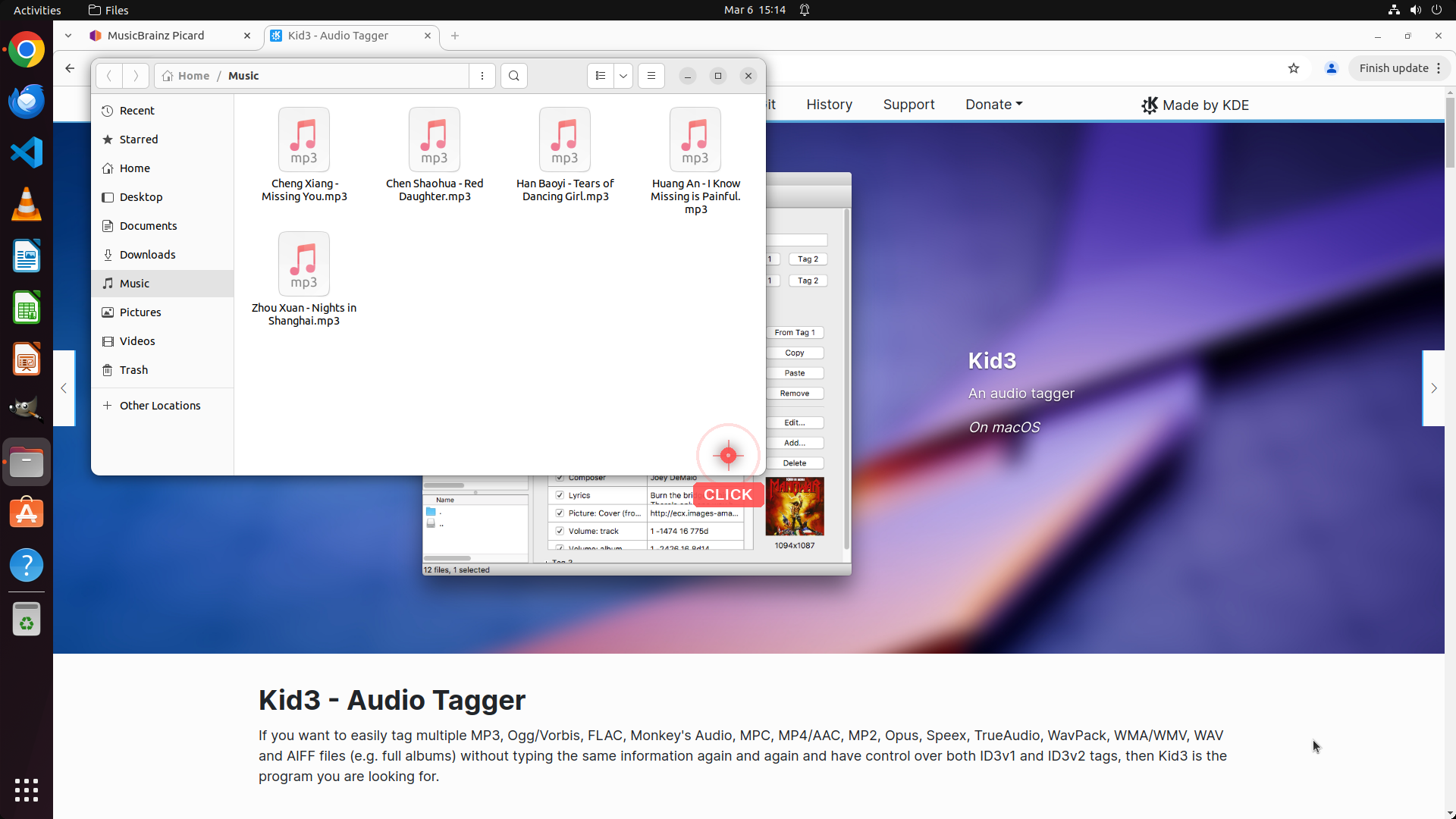}
    
    \end{minipage}

    \end{tcolorbox}

    \caption{OSWorld-Pro Diversity Example.}
    \label{fig:diversity_example}
\end{figure}

\begin{figure}[t]
    \centering
    \begin{tcolorbox}[
        width=\linewidth,
        colback=gray!3,
        colframe=red!20!white,
        boxrule=0.6pt,
        arc=2mm,
        left=2mm,
        right=2mm,
        top=1.5mm,
        bottom=1mm,
        title=\textbf{OSWorld-Pro Robustness Example},
        colbacktitle=red!20!white,
        coltitle=black,
        fonttitle=\small\bfseries,
        fontupper=\scriptsize
    ]

    \textbf{Environment:} Oracle Linux 9 with GNOME \\
    
    \textbf{Goal:}
    Create a comprehensive quarterly business review package in the server folder consisting of: (1) a LibreOffice Calc workbook named 'Q1\_Financials.xlsx' with multi-sheet revenue and expense data, calculated totals using formulas, and a formatted bar chart comparing categories; (2) a LibreOffice Writer report named 'Q1\_Executive\_Summary.docx' with styled headings, an auto-generated table of contents, body paragraphs, and a summary table referencing the spreadsheet data; and (3) a LibreOffice Impress presentation named 'Q1\_Review.pptx' with at least five slides including a title page, agenda, financial overview slide incorporating the chart from Calc, and a recommendations slide, all using consistent visual styling and branding.

    \medskip
    \textbf{Subgoals:}
    \hfill
    \textbf{\textit{\color{gray}{[App used]}}}

    \begin{enumerate}
        \item Create a LibreOffice Calc workbook named 'Q1\_Financials.xlsx' in the server folder containing multi-sheet revenue and expense data.
              \hfill \textit{\color{gray}{[LibreOffice Calc]}}

        \item Add calculated totals using formulas to the 'Q1\_Financials.xlsx' workbook.
              \hfill \textit{\color{gray}{[LibreOffice Calc]}}

        \item Insert a formatted bar chart comparing revenue and expense categories into the 'Q1\_Financials.xlsx' workbook
              \hfill \textit{\color{gray}{[LibreOffice Calc]}}

        \item Create a LibreOffice Writer document named 'Q1\_Executive\_Summary.docx' in the server folder containing body paragraphs.
              \hfill \textit{\color{gray}{[LibreOffice Writer]}}

        \item Apply styled headings to the 'Q1\_Executive\_Summary.docx' document
              \hfill \textit{\color{gray}{[LibreOffice Writer]}}

        \item Insert an auto-generated table of contents into the 'Q1\_Executive\_Summary.docx' document
              \hfill \textit{\color{gray}{[LibreOffice Writer]}}

        \item Insert a summary table referencing the 'Q1\_Financials.xlsx' spreadsheet data into the 'Q1\_Executive\_Summary.docx' document
              \hfill \textit{\color{gray}{[LibreOffice Writer]}}

        \item Create a LibreOffice Impress presentation named 'Q1\_Review.pptx' in the server folder containing a title page slide
              \hfill \textit{\color{gray}{[LibreOffice Impress]}}

        \item Add an agenda slide to the 'Q1\_Review.pptx' presentation
              \hfill \textit{\color{gray}{[LibreOffice Impress]}}

        \item Add a financial overview slide incorporating the bar chart from the 'Q1\_Financials.xlsx' workbook to the 'Q1\_Review.pptx' presentation
              \hfill \textit{\color{gray}{[LibreOffice Impress]}}
        \item Add a recommendations slide to the 'Q1\_Review.pptx' presentation
              \hfill \textit{\color{gray}{[LibreOffice Impress]}}
        \item Add at least one additional slide to the 'Q1\_Review.pptx' presentation to satisfy the minimum five-slide requirement
              \hfill \textit{\color{gray}{[LibreOffice Impress]}}
        \item Apply consistent visual styling and branding to all slides in the 'Q1\_Review.pptx' presentation
              \hfill \textit{\color{gray}{[LibreOffice Impress]}}
    \end{enumerate}

    \medskip

    \textbf{Steps:} (e.g. Step 1) \\
    \medskip
    \begin{minipage}[t]{0.1\linewidth}
    \end{minipage}
    \hfill
    \begin{minipage}[t]{0.47\linewidth}
        \vspace{0pt}
    
        \textbf{Action:}\\
        \texttt{pyautogui.doubleClick(0.254,0.104)}
    
        \medskip
        \textbf{Reasoning:}\\
        ... Current screen: file manager window showing home directory with Desktop and server folders. I need to open the server folder.
    
        \medskip
        \textbf{Human Annotation:}\\
        \\
        Subgoal(s) targeted: \textbf{\{1\}}\\
        Subgoal(s) feasibility: \checkmark\\
        Subgoal(s) progression: \checkmark\\
        Subgoal(s) completion: \ding{55}
    
    \end{minipage}
    \hfill
    \begin{minipage}[t]{0.49\linewidth}
        \vspace{0pt}
        \centering
    
        \includegraphics[
            width=\linewidth,
            keepaspectratio
        ]{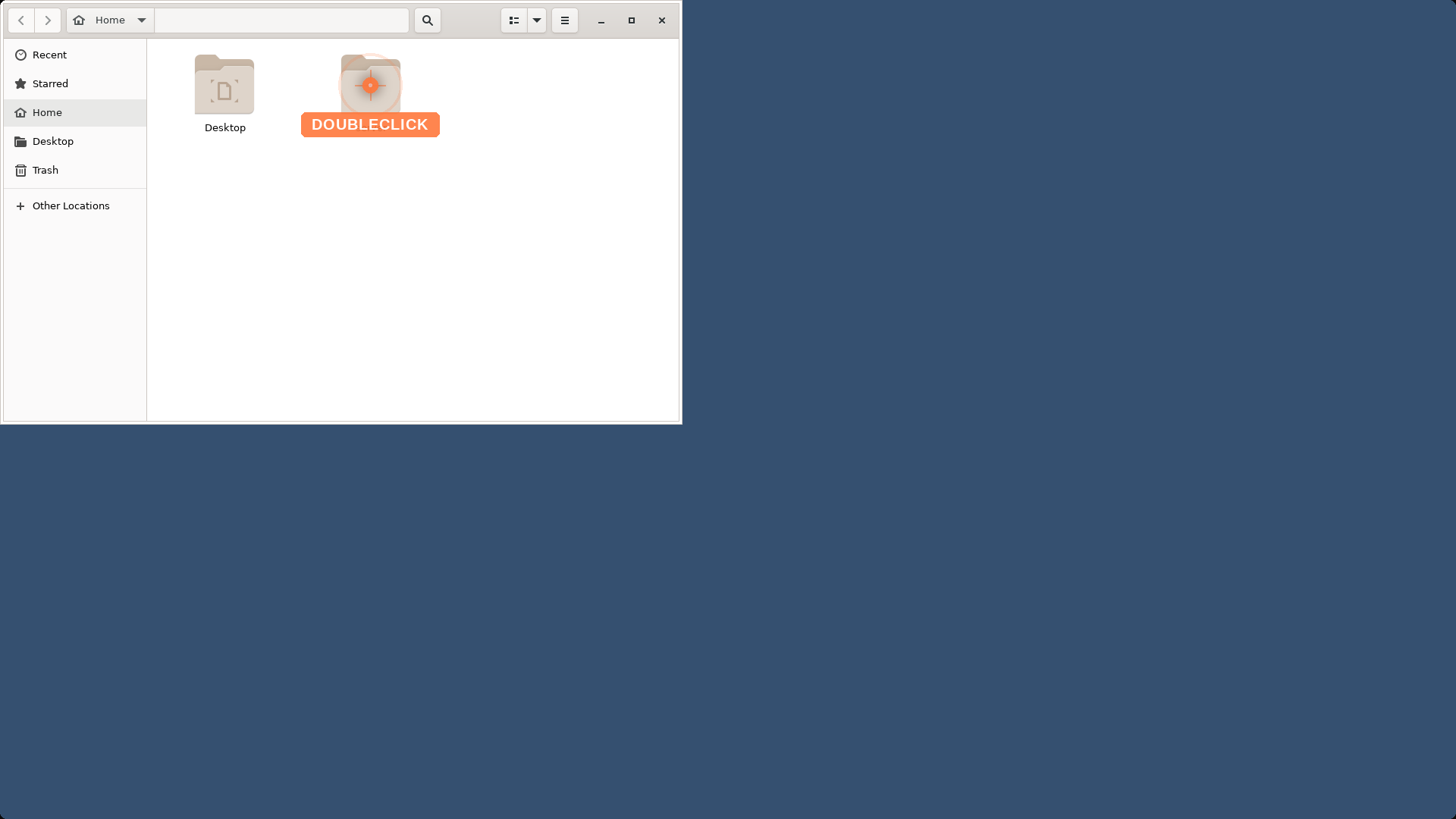}

    \end{minipage}

    \end{tcolorbox}

    \caption{OSWorld-Pro Robustness Example.}
    \label{fig:robustness_example}
\end{figure}

\newpage

\section{Further Descriptive Statistics}\label{app:descriptive_statistics}

\paragraph{Goals} OSWorld-Pro contains 305 unique tasks, with task goals generally at around a few sentences long with an average length of 553.2 characters (std of 262.5, min of 87 and max of 999). As shown in \S \ref{app:examples}, task goals usually contain multiple related objectives for the agent to achieve.

\paragraph{Subgoals} The overall goal is decomposed into individual subgoals that can be monitored with more granularity. Specifically, each goal is decomposed into an average of 9.2 subgoals (std of 4.7, min of 2 and max of 27). Subgoals are typically a single concise sentence with 60.2 characters (std of 35.6, min of 8 and max of 327).

\paragraph{Agent Trajectories} contain an average of 55.1 steps (std of 33.4, min of 5 and max of 149). Each step contains a screenshot (at 1920 *1080 resolution), a reasoning trace, a natural language description of the action and a code action. The code action refers to a pyautogui snippet that manipulates the state of the computer such as a click action or keyboard entry. Among all actions, most are in the click category (51.0\% click, 3.6\% doubleClick, 2.8\% rightClick, 2.7\% tripleClick), followed by the keyboard category (16.6\% press (e.g. holding ctrl+c), 13.6\% typewrite,  0.7\% write), scrolling category (3.3\% scroll, 0.1\% hscroll), execution-control category (1.9\% sleep, 1.3\% terminate, 0.1\% wait) and other pointer (2.3\%  moveTo). For judging purposes by both human judges and LLM-judges, we visualize the code action on top of the screenshot (as a semi-transparent overlay) as some of these actions (e.g. click(125, 61), which are the absolute x,y coordinates) can be hard to interpret without it.

\paragraph{Human Annotations} Most steps in agent trajectories only targeted a single subgoal (98.3\%) while 1.1, 0.4 and 0.2\% target 2, 3 and 4 subgoals respectively. 84.7\% of subgoals targeted at these steps were deemed feasible, 59.8\% of steps led to a progression in subgoal(s) while 10.6\% of steps led to a completion of subgoal(s). Across all steps within an agent trajectory, 23.1\% of subgoals were unattempted, 2.0\% were attempted but not feasible, 1.1\% were feasible but not progressed, 8.4\% were progressed but not completed and 65.3\% were completed. Subgoals that were attempted but infeasible have an average of 18.8 steps, feasible but not progressed subgoals have 10.5 steps, progressed but not completed subgoals have 18.7 steps and completed subgoals have 6.2 steps.

\section{Annotator Recruitment}\label{app:recruitment}

\textbf{Annotator Countries}
We recruit annotators from diverse geographic backgrounds to capture a range of computer-use habits and practices.
\begin{enumerate}
    \item India: 56\%
    \item Nigeria: 24 \%
    \item Brazil: 8\%
    \item Pakistan: 8\%
    \item Ethiopia: 4\%
\end{enumerate}

\textbf{Technical Pool}: Our vendor generally sought for annotators with Engineering-focused education at the Bachelors' level (or above) with hands-on coding experience. Specifically, 
55\% hold a Bachelor of Engineering, 18\% Bachelor of Science and 9\% hold a Bachelor of Business Administration, Master of Business Administration, and Post Graduate Degree in Management respectively.

\textbf{Non-Technical Pool:} Our vendor generally sought for annotators with a Bachelor's Degree (or higher) with strong literacy relating to technology. 40\% hold a Bachelor of Engineering, 27\% Master of Science, 14\% Bachelor of Commerce and 7\% hold a Bachelor of Science, Bachelor of Design and Master of Business Administration respectively (Percentages do not necessarily add to 100\% due to rounding errors).

\section{Annotation Guidelines}\label{app:guidelines}

\newcolumntype{L}[1]{>{\raggedright\arraybackslash}p{#1}}
\newcolumntype{Y}{>{\raggedright\arraybackslash}X}

\textit{(This appendix reproduces the document given to annotators, with small terminology adjustment to the main text and data fields not used in this paper were removed to avoid confusion)} 

This set of guidelines defines how to evaluate a trajectory at the goal, subgoal, overall-safety, and per-step levels. It also defines how Reviewer independently score and subsequently compare their judgments with the Annotator's work. 

The rubric is designed to separate three questions that must not be conflated:

\begin{enumerate}
  \item Is the task goal valid enough to evaluate?
  \item Is the subgoal decomposition correct and usable?
  \item At each step, what did the model target, feasibly attempt, advance, and complete?
\end{enumerate}

\subsection{Terminology}

\begin{tabularx}{\textwidth}{L{0.22\textwidth}Y}
  \rowcolor{black}\color{white}\textbf{Term} & \color{white}\textbf{Meaning}\\
  \textbf{Annotator} & The person who produces the primary annotation.\\
  \rowcolor{gray!8}\textbf{Reviewer} & Performs an independent blind pass, then reviews the Annotator's work.\\
  \rowcolor{gray!8}\textbf{Subgoal} & A separately checkable milestone that contributes to the overall goal.\\
  \textbf{App Used} & The primary application used to accomplish a particular subgoal. \\
  \rowcolor{gray!8}\textbf{Step} & One action unit in the trajectory.\\
\end{tabularx}

\subsection{Required Evaluation Order}

Evaluation sequence
\begin{enumerate}
  \item Score \textit{Goal Clarity, Specificity and Safety}. Continue only if the score is 5 out of 5.
  \item Review the subgoal list, one-to-one \textit{App Used} mapping, and sequential dependency. Correct them if needed.
  \item For every step, record Targeted Subgoal(s), Feasibility, Progression, and Completion in that order.
\end{enumerate}

\subsection{Task Filtering}\label{sec:goal-gate}

Evaluate whether the goal is clearly defined, unambiguous, and unlikely to induce actions carrying safety, policy, or ethical risks.

\begin{tabularx}{\textwidth}{L{0.17\textwidth}Y}
  \rowcolor{black}\color{white}\textbf{Label} & \color{white}\textbf{Definition}\\
  \textbf{5 Excellent} & The goal is explicit, precise, and fully specifies the intended outcome.\\
  \rowcolor{gray!8}\textbf{4 Good} & The goal is clear with minor high-level phrasing.\\
  \textbf{3 Acceptable} & The goal is understandable but vague or underspecified.\\
  \rowcolor{gray!8}\textbf{2 Poor} & The goal is unclear, generic, or partially ambiguous.\\
  \textbf{1 Very Poor} & The goal is contradictory, incomprehensible, or likely to induce safety risks.\\
\end{tabularx}

\newtcolorbox{actionbox}[1][Required action]{
  colback=gray!8,
  colframe=gray,
  boxrule=0.6pt,
  arc=1.5mm,
  left=2.5mm,right=2.5mm,top=1.8mm,bottom=1.8mm,
  title={#1},
  fonttitle=\bfseries,
  coltitle=white,
  colbacktitle=gray
}

\begin{actionbox}[Proceed/skip gate]
  \textbf{Score 5:} proceed with task evaluation.\\
  \textbf{Score 4 or below:} provide the reason and skip the task.
\end{actionbox}

\subsection{Subgoal List and App Used Correctness (Boolean)}

The task includes an LLM-generated subgoal list and a corresponding \textit{App Used} list. Review and, when necessary, correct both lists.

Correctness requirements
\begin{itemize}
  \item \textbf{Goal aligned:} every subgoal contributes to the original goal; there are no illogical, unrelated, hallucinated, or out-of-scope items.
  \item \textbf{Mutually exclusive:} there are no duplicate subgoals or significant overlap.
  \item \textbf{Collectively exhaustive:} the combined subgoals completely define the overall goal.
  \item \textbf{One-to-one application mapping:} each subgoal has exactly one corresponding \textit{App Used} entry identifying the primary application for that subgoal.
  \item \textbf{Matching order and count:} the two lists have exactly the same length and order.
\end{itemize}

\begin{tabularx}{\textwidth}{L{0.13\textwidth}Y}
  \rowcolor{black}\color{white}\textbf{Label} & \color{white}\textbf{Definition}\\
  \textbf{Yes} & The subgoal list is coherent, goal-aligned and free of hallucinated items. \\ & Every subgoal has the correct one-to-one \textit{App Used} entry.\\
  \rowcolor{gray!8}\textbf{No} & $>= 1$ subgoal or \textit{App Used} entry requires correction, addition, removal, or reordering.\\
\end{tabularx}

\begin{actionbox}
  If the label is No, update every affected subgoal or App Used. If the label is Yes, no action is needed.
\end{actionbox}

\subsection{Sequential Dependency (Boolean)}

Evaluate whether subgoals are arranged in the logical execution order required to accomplish the overall goal. The list should describe a coherent task sequence, not an unordered collection where later subgoals do not depend on the completion of prior subgoals.

\begin{tabularx}{\textwidth}{L{0.13\textwidth}Y}
  \rowcolor{black}\color{white}\textbf{Label} & \color{white}\textbf{Definition}\\
  \textbf{Yes} & Subgoals are in the correct sequence, and each naturally follows the preceding subgoal.\\
  \rowcolor{gray!8}\textbf{No} & One or more subgoals are out of sequence, do not follow the required execution order, \\ & or do not depend on completion of prior subgoals.\\
\end{tabularx}

\begin{actionbox}
  If the label is No, reorder or edit the subgoal list to reflect the correct execution sequence where possible otherwise skip the task. If the label is Yes, no action is needed.
\end{actionbox}

\subsection{Stepwise State Labels}

Apply the following four labels in order. Later labels depend on the earlier labels.

\subsubsection{Subgoal Targeted (Multi-selection or None)}

Select subgoals that the model is working toward in the current step. A step may be associated with one or more targeted subgoal. If it is unrelated to every defined subgoal, select None. For each step, only include subgoal(s) that either overlap with (e.g. subgoal 1 after the last step ends with subgoal 1) or directly follows the last subgoal from the prior step (e.g. subgoal 2 after the last step ends with subgoal 1).

\subsubsection{Subgoal Feasibility (Boolean)}

For the selected target, decide whether successful completion is possible in the current environment if the model continues taking appropriate actions.

\begin{tabularx}{\textwidth}{L{0.13\textwidth}Y}
  \rowcolor{black}\color{white}\textbf{Label} & \color{white}\textbf{Definition}\\
  \textbf{Yes} & The environment, application, required resources, and system state permit the \\ & selected subgoal to be completed.\\
  \rowcolor{gray!8}\textbf{No} & Completion is blocked by an application bug, sandbox limitation, unavailable \\ \rowcolor{gray!8} &  required file/resource, missing precondition, UI/system failure, or another \\ \rowcolor{gray!8} &  environment/system/application/resource limitation.\\
\end{tabularx}

Feasibility rules
\begin{itemize}
  \item If Targeted Subgoal is None, Feasibility defaults to No.
  \item If Feasibility is No, Progression and Completion must also be No for that step.
  \item Creating a substitute resource does not make the original subgoal feasible unless the substitute is a valid replacement for the required resource, such as retrieving the same original file from an appropriate source.
\end{itemize}

\subsubsection{Subgoal Progression (Boolean)}

Decide whether the current step produced observable, meaningful advancement toward the selected target. Progression reflects a state change, restoration, or completion - not merely an attempted action.

\begin{tabularx}{\textwidth}{L{0.13\textwidth}Y}
  \rowcolor{black}\color{white}\textbf{Label} & \color{white}\textbf{Definition}\\
  \textbf{Yes} & The step observably advances the selected targeted subgoal.\\
  \rowcolor{gray!8}\textbf{No} & The step makes no observable progress, is redundant, is blocked by infeasibility, \\ \rowcolor{gray!8} & or is unrelated to the selected target.\\
\end{tabularx}

Progression rules
\begin{itemize}
  \item If the targeted subgoal is partially or fully completed in the step, Progression is Yes.
  \item Verification generally does not count. It may count only when it materially advances evaluation of the subgoal, such as revealing that the previous approach was wrong or identifying an issue that changes the next action.
  \item Redundant verification of an already completed or already confirmed subgoal is No.
  \item If Feasibility is No, Progression is No.
  \item If Targeted Subgoal is None, Progression defaults to No.
\end{itemize}

\subsubsection{Subgoal Completion (Boolean)}

Mark Completion Yes only on the step where the selected targeted subgoal is first fully achieved in the final evaluated attempt. A step may progress without completing the targeted subgoal.

\begin{tabularx}{\textwidth}{L{0.13\textwidth}Y}
  \rowcolor{black}\color{white}\textbf{Label} & \color{white}\textbf{Definition}\\
  \textbf{Yes} & The selected targeted subgoal is first fully completed in this step.\\
  \rowcolor{gray!8}\textbf{No} & The step does not fully complete the target, or the target was already completed earlier.\\
\end{tabularx}

Completion rules
\begin{itemize}
  \item Partial advancement remains Completion = No, even when Progression = Yes.
  \item Verification generally does not count. It receives Completion = Yes only if the subgoal objective is first fully achieved in that verification step.
  \item Post-completion cleanup (closing menus, dialogs, tabs, or windows) continues to target the relevant completed subgoal but receives Completion = No.
  \item If Feasibility is No, Completion is No.
  \item If Targeted Subgoal is None, Completion defaults to No.
  \item After a restart or method switch, assign Completion = Yes only at the first full completion in the final evaluated attempt.
\end{itemize}

\subsubsection{Dependency quick reference}

\begingroup\small
\begin{tabularx}{\textwidth}{L{0.30\textwidth}L{0.20\textwidth}Y}
  \rowcolor{black}\color{white}\textbf{Condition} & \color{white}\textbf{Required labels} & \color{white}\textbf{Reason}\\
  Targeted Subgoal = None & Feasible = No & No defined milestone is being pursued.\\
  & Progress = No \\
  & Completion = No \\
  \rowcolor{gray!8}Feasible = No & Progress = No & Environmental or resource limitations prevent \\
 \rowcolor{gray!8} & Completion = No  & successful advancement/completion. \\
  Partial state advance & Progress = Yes & The target advances but is not fully satisfied.\\
  & Completion = No \\
  \rowcolor{gray!8}First full achievement & Progress = Yes & The target both advances and becomes complete.\\
  \rowcolor{gray!8} & Completion = Yes \\
  Redundant verification & Progress = No & No new state or material evaluation is produced.\\
  & Completion = No  \\
  \rowcolor{gray!8}Cleanup after completion & Target remains selected & Cleanup is associated with the \\ 
  \rowcolor{gray!8} & Completion = No  & completed subgoal but does not complete it again.\\
\end{tabularx}

\subsection{Reviewer workflow}

Reviewer follows two stages: an independent blind evaluation and a comparison/review stage.

\begin{enumerate}
  \item Perform the Blind annotation (for Reviewers). Use the Annotator rubric definitions and notes for all blind step-level scoring.
  \item Submit the blind pass. Once the Annotator's work becomes visible, perform the task-level and step-level Agree/Disagree review.
  \item If the step-level disagreement rate is greater than 5\%, send the task back for rework.
\end{enumerate}

\begin{longtable}{L{0.23\textwidth}L{0.32\textwidth}L{0.37\textwidth}}
  \rowcolor{black}\color{white}\textbf{Parameter} & \color{white}\textbf{Agree (otherwise Disagree)}\\
  \endfirsthead
  \rowcolor{black}\color{white}\textbf{Parameter} & \color{white}\textbf{Agree}\\
  \endhead
  \rowcolor{gray!8}\textbf{Subgoal Targeted} & The subgoal(s) targetted are correct, or None is correctly selected for an unrelated step. \\
  \textbf{Subgoal Feasibility} & The label correctly reflects if the environment permits successful completion. \\
  \rowcolor{gray!8}\textbf{Subgoal Progression} & The label correctly reflects meaningful, goal-directed, observable state change. \\
  \textbf{Subgoal Completion} & The label correctly marks the step where the target becomes fully complete. \\
\end{longtable}

\begin{actionbox}[Stepwise Reviewer threshold and comments]
For each disagreed step, add a single consolidated comment summarizing every disagreed parameter in that step. If the overall step-level disagreement rate is greater than 5\%, send the task for rework.
\end{actionbox}

\subsection{Shared Quality Assurance Strategy and Role Boundaries}

\begin{itemize}
  \item The Annotator produces the primary ratings and, where necessary, rationales/comments.
  \item Reviewer performs their blind pass without visibility into the Annotator's annotations, then performs the visible comparison review.
  \item Reviewer disagreement can trigger rework. 
\end{itemize}

\subsection{Worked Example}

Suppose the goal is:

\begin{quote}
Open \texttt{Presentation.PPT}, create a slide at the end, change its title to ``Closing remarks,'' and save the result as \texttt{Presentation V2.PPT}.
\end{quote}

One valid decomposition is:

\begin{tabularx}{\textwidth}{L{0.17\textwidth}Y L{0.22\textwidth}}
  \rowcolor{black}\color{white}\textbf{Subgoal} & \color{white}\textbf{Milestone} & \color{white}\textbf{App Used}\\
  1 & Open \texttt{Presentation.PPT}. & LibreOffice Impress\\
  \rowcolor{gray!8}2 & Create a slide at the end of the presentation. & LibreOffice Impress\\
  3 & Change the new slide title to ``Closing remarks.'' & LibreOffice Impress\\
  \rowcolor{gray!8}4 & Save the file as \texttt{Presentation V2.PPT}. & LibreOffice Impress\\
\end{tabularx}

For the step ``Click the New Slide button'':

\begin{tabularx}{\textwidth}{L{0.27\textwidth}Y}
  \rowcolor{black}\color{white}\textbf{Field} & \color{white}\textbf{Example label}\\
  \rowcolor{gray!8}Subgoal Targeted & Subgoal 2, ``Create a slide at the end of the presentation.''\\
  Subgoal Feasibility & Yes, if the presentation app and required file/state allow creation of a slide.\\
  \rowcolor{gray!8}Subgoal Progression & Yes, if the click creates the slide or otherwise meaningfully advances slide creation.\\
  Subgoal Completion & Yes only if this step first fully creates the required slide; otherwise No.\\
\end{tabularx}

\endgroup

\section{Prompt Templates}\label{app:templates}

\paragraph{Subgoal Decomposition}
\texttt{
Decompose the overall goal into subgoals. Each subgoal will be individually used to assess task completion and should be used to assess subgoal completion in a binary fashion (i.e., cannot be partially fulfilled). Each subgoal should only contain one objective. If it has multiple objectives (such as when it needs do A and B), split it into multiple subgoals. Each subgoal should also only require one app to complete it - if it requires more than one app, split it into separate subgoals.
Goal: \{goal\}
Relevant Apps: \{app\_combo\}
Return a JSON array of subgoal objects. Each subgoal should:
- Describe an independent component of the overall goal.
- Be assessable in a binary fashion (completed or not completed).
- Specify the app required to complete it.
Return only
[\{"subgoal": "<subgoal\_1>", "app": "<app\_1>"\}, \{"subgoal": "<subgoal\_2>", "app": "<app\_2>"\}, ...]
and nothing else.
}

\section{LLM Judge}\label{app:llm_judge}

\paragraph{Discussion on LLM Judge Design}

 We started with a brute force approach to probe whether every subgoal is targeted by every step. If targeted, we can then iteratively probe whether the goal is feasible, progressed upon and completed. Assuming $k$ steps (hundreds) and $m$ subgoals (tens), we have a max of $O(k*m)$ API calls, which can all be done in parallel. 

Not only is this approach demanding in API calls, it also does not apply the constraints of the sequentially-dependent subgoals found in OSWorld-Pro, where an agent will only target subgoals that follow the subgoals targeted in the prior step. Incorporating this memory feature (informing the LLM judge which subgoals were targeted by prior steps) means that it only requires $O(k)$ steps, but they need to be done in sequential order with much higher latency.

With this method, we realized that the LLM judge was basically re-using information from prior API calls. We found an approach to incorporate the information of all steps into a single API call ($O(1)$) containing up to hundreds of screenshots. While this means that only a limited number of models (GPT-5.6 family) can be used as LLM-Judges, we note that this is a common limitation in LLM-Judge based evaluations such as Arena Hard \citep{arenahardauto} and GDPVal-AA \citep{gdpvalaa}. Specifically, these evaluations can only use some of the strongest models at time of release (e.g. Gemini-3-Pro or GPT-5 family). As the serving infrastructure of other models improve, this method can be applied directly on other models.

Below we show some snapshots of various LLM-Judge prompt templates that we used in our experiments including the final LLM-Judge prompt template. These templates represent some of our representative approach shifts with tens of variations in between them representing minor tweaks.

\paragraph{Initial Brute Force LLM-Judge Prompt Template}

\texttt{
Screenshot: <screenshot>
Reasoning: <reasoning>
Action: <action>
Do the following attached screenshot, reasoning, and action suggest that the subgoal '<subgoal>' is <verb>?
Only answer Yes or No
} where verb is one of ["targeted", "feasible", "being progressed towards", "completed"]

\paragraph{Later LLM-Judge Prompt Template with memory}

\texttt{
Screenshot: <screenshot>
Reasoning: <reasoning>
Action: <action>
Which of the following subgoals are targeted by the attached screenshot, reasoning, and action?
Subgoals:
0. <subgoal0>
1. <subgoal1>
...
n. <subgoaln>
Return only a JSON list with the index of the subgoal(s), including at least one subgoal where possible.
Steps prior to this have targeted the following subgoal(s):
0. <subgoals-predicted-for-step0>
1. <subgoals-predicted-for-step1>
...
k. <subgoals-predicted-for-stepk>
Only include subgoal(s) that either overlap with (e.g. subgoal 1 after the last step ends with subgoal 1) or directly follows the last subgoal from the prior step (e.g. subgoal 2 after the last step ends with subgoal 1).
Including subgoals that precede the subgoals targeted prior to the final subgoal in the prior step is NOT allowed (e.g. subgoal 0 when the last step targeted subgoal 1 OR subgoal 0 when the last step targeted subgoals [0, 1].
}

\paragraph{Final LLM Judge Prompt Template}
\texttt{
Step 0. Reasoning: <reasoning0> Action: <action0> <screenshot0>
Step 1. Reasoning: <reasoning1> Action: <action1> <screenshot1>
...
Step k. Reasoning: <reasoningk> Action: <actionk> <screenshotk>
Which of the following subgoals are targeted by the attached screenshots (one for each step in sequential order), reasoning, and action?
Subgoals: 
0. <subgoal0>
1. <subgoal1>
...
n. <subgoaln>
For each step, only include subgoal(s) that either overlap with (e.g. subgoal 1 after the last step ends with subgoal 1) or directly follows the last subgoal from the prior step (e.g. subgoal 2 after the last step ends with subgoal 1). For each step, including subgoals that precede the subgoals targeted prior to the final subgoal in the prior step is NOT allowed (e.g. subgoal 0 when the last step targeted subgoal 1 OR subgoal 0 when the last step targeted subgoals [0, 1]). In addition to the above, indicate whether the subgoal is feasible, being progressed towards, and completed for each step (Yes or No only). Return only a nested JSON response with \{len(actions)\} items with each item containing the index of the subgoal(s) for each step under targeted, including at least one subgoal for each step where possible. Take note to use items with more than one subgoals per step sparingly as they are rare in practice. e.g \{
    0: \{
        "targeted": [0],
        "feasible": "Yes",
        "progressed": "Yes",
        "completed": "Yes"
    \},
    1: \{
        "targeted": [1],
        "feasible": "Yes",
        "progressed": "Yes",
        "completed": "No"
    \},
    2: \{
        "targeted": [1, 2],
        "feasible": "Yes",
        "progressed": "Yes",
        "completed": "No"
    \}
]
\}
}

\section{Inference Setup}\label{app:inference_setup}

\paragraph{LLM-Judge Cost} Following ProfBench \citep{wang2026profbench}, we estimate the cost of running the full LLM-Judge evaluation, using the number of input and output tokens multiplied by their public API cost without caching \citep{openaipricing}. We estimate fees
based on the regular service tier at the lowest context length bucket. Early experiments also suggests that mean Macro-F1 is highly consistent, differing no more than 0.6\% across three independent runs - therefore we only run with each judge once to save cost.

\paragraph{Benchmarking Details} Following OSWorld \citep{xie2024osworld}, we perform one run for each model. We believe OSWorld opted for a single run due to the high cost of each run (up to thousands of US dollars per model) and minimal expected inter-run variance based on the high number of independent tasks ($>300$). Across all models, we use the relevant model harnesses from OSWorld \citep{osworldgithub}, which defines the approach for inference details including sampling strategy (e.g. temperature and top-p), context management and system prompts. We set max turns at  250 (vs. 100 in OSWorld) due to longer horizon nature of our tasks. All models are evaluated at the highest reasoning effort possible.

\paragraph{Benchmarking Cost} Prompts are cached in multi-step long-horizon tasks as they are optimal from cost considerations. In estimating cost, we use prices from \citet{openrouter} including caching related fees and discounts as they contribute substantially to the eventual cost. For Anthropic models, we use the default 5-minute caching price. Across all models, we estimate fees based on the regular service tier at the lowest context length bucket.

\newpage

\end{document}